\documentclass[letterpaper, 10pt]{article}
\usepackage[margin=2.5cm]{geometry}

\usepackage[usenames, dvipsnames]{color}
\usepackage[noadjust]{cite}
\usepackage{booktabs}
\usepackage{graphicx}
\usepackage{caption}
\usepackage{subcaption}
\usepackage{float}
\usepackage{amsmath}
\usepackage{algorithmicx}
\usepackage{subfiles}
\usepackage{multirow}
\usepackage{mathrsfs}

\usepackage[all,defaultlines=4]{nowidow}

\graphicspath{{images}{../images/}}

\usepackage[usenames,dvipsnames,table]{xcolor}
\usepackage{soul}
\usepackage[export]{adjustbox}

\usepackage[pdfencoding=auto, pdfusetitle,hidelinks]{hyperref}

\title{\LARGE \bf
Building Prior Knowledge: A Markov Based Pedestrian Prediction Model Using Urban Environmental Data
}

\hypersetup{
  pdftitle={Building Prior Knowledge: A Markov Based Pedestrian Prediction Model Using Urban Environmental Data},
  pdfauthor={Pavan Vasishta, Dominique Vaufreydaz, Anne Spalanzani},
  pdfkeywords={Situational Awareness, Pedestrian Behaviour, Hidden Markov Models, Autonomous Vehicles.},
}

\author{Pavan Vasishta$^{1}$, Dominique Vaufreydaz$^{2}$ and Anne Spalanzani$^{1}$}

\date{
\small{\textsuperscript{1}~Univ. Grenoble Alpes, Inria, 38000 Grenoble, France\\
\textsuperscript{2}~Univ. Grenoble Alpes, CNRS, Inria,  Grenoble INP, LIG, 38000 Grenoble France}\\
\vspace{0.5em}\textit{Author version}
}

\begin{document}

\maketitle
\thispagestyle{empty}
\pagestyle{empty}
\setlength{\belowdisplayskip}{2pt}
\widowpenalty10000
\clubpenalty10000
\addtolength{\abovedisplayskip}{-5pt}

\begin{abstract}

Autonomous Vehicles navigating in urban areas have a need to understand and predict future pedestrian behavior for safer navigation. This high level of situational awareness requires observing pedestrian behavior and extrapolating their positions to know future positions. While some work has been done in this field using Hidden Markov Models (HMMs), one of the few observed drawbacks of the method is the need for informed priors for learning behavior. In this work, an extension to the Growing Hidden Markov Model (GHMM) method is proposed to solve some of these drawbacks. This is achieved by building on existing work using potential cost maps and the principle of \textit{Natural Vision}. As a consequence, the proposed model is able to predict pedestrian positions more precisely over a longer horizon compared to the state of the art. The method is tested over ``legal'' and ``illegal'' behavior of pedestrians, having trained the model with sparse observations and partial trajectories. The method, with no training data, is compared against a trained state of the art model. It is observed that the proposed method is robust even in new, previously unseen areas.

\end{abstract}

\vspace{0.5em}
\begin{changemargin}{0.9cm}{0.9cm} 
\end{changemargin}
\vspace{0.5em}

\section{INTRODUCTION} \label{sec:introduction}

Autonomous vehicles are slowly seeping into the popular consciousness. With the commercial deployment of self driving cars imminent, concerns about safety have taken center stage. More so regarding their viability in urban streets and urban centers. In urban centers, autonomous vehicles face scenarios such as navigating between unruly drivers, share space with cyclists, bikers and pedestrians. These cars need to observe these shared space users and predict their behavior. To do so, a high level of Situational Awareness \cite{Endsley1995} needs to be anticipated. This required level of Situational Awareness deals with comprehending the different elements of the scene and projecting the future positions of these elements.

Understanding pedestrian behavior in static scenes has been widely explored using learning methods. Observed trajectories are used to build a cost function of the observed environment. Then Markov Decision Processes are used to solve the pedestrian trajectory problem \cite{Ziebart2009}. Work done in \cite{Kitani2012} deals with knowledge transfer of the cost function based on the semantic labeling of the observed scene which then uses Partially Observed Markov Decision Processes (POMDPs) to predict pedestrian positions. 

In \cite{Bandyopadhyay2013}, a pedestrian's intended destination is modeled as a hidden variable and solves a POMDP to navigate an autonomous vehicle in crowded areas. Another method for goal estimation can be found in \cite{Yamaguchi2011} which minimizes an energy function that is used to infer intended positions based on previous observations. The principle of using goals as a part of the state space using Hidden Markov Models (HMMs) can also be found in \cite{Vasquez2009}. Here, the authors describe a continuous learning algorithm called a Growing Hidden Markov Model (GHMM) wherein the number of states of the HMM are changing over time. A drawback of this model is that it requires knowledge of the complete trajectory sequence of the pedestrian before any inference is performed. That is to say, the starting and ending points of all the training trajectories need to be known beforehand making it difficult to use in scenarios where observed pedestrians might be temporarily occluded or lost to the tracker. 

To solve this issue, an extension to the GHMM method was proposed \cite{Hurtado2015} which deals with automatic identification of goals through partial trajectory observation. Both these implementations of the GHMM require a large dataset of trajectories in the environment to accurately model, and thus infer future pedestrian behavior. A secondary extension to the GHMM method was proposed in \cite{Elfring2014} to incorporate motion models and social forces \cite{Helbing1995} for a more accurate pedestrian goal prediction.

In a previous work, a new method to model an urban environment, based on the sociological principle of \textit{Natural Vision} \cite{Gibson1979}, had been introduced. This principle suggests that pedestrian motion in an environment is decided by certain attractors in it, called Points of Interest (POIs). The current work utilizes these POIs to generate potential cost maps of the environment which are also dependent on its semantics, i.e, the sidewalk, crosswalk, width of the road, etc. The main motivation of this work is to build upon those ideas to predict pedestrian behavior in environments with sparse observations, or in completely new, previously unobserved areas. 
Here, an extension is proposed to the GHMM method that initializes a ``Prior Topology'' allowing for generation of more accurate topological connections for the environment to describe pedestrian paths. This extension also leads to a more realistic initialization of the HMM parameters, allowing for quick model convergence and correspondingly, better inference. The extension has been designed to be used on an autonomous car. The quick convergence on fewer partial trajectories, compared to the state of the art, implies that it can be used on an autonomous platform to predict pedestrian behavior even in previously unseen, dense urban environments. 

This paper has been divided as follows: Section \ref{sec:primer} acts as a short primer to the state-of-the-art in GHMMs and its shortcomings. It also discusses briefly, the previously developed method to generate a potential cost map based on the observed environment \cite{Vasishta2017}. Section \ref{sec:approach} explains the extension to the GHMM algorithm. Section \ref{sec:implementation} discusses the implementation and the results of the method applied to an available dataset. Finally, Section \ref{sec:conclusion} discusses conclusions and future work of the current approach.  

\section{Related Work} \label{sec:primer}

\subsection{Growing Hidden Markov Models}

A GHMM is a discrete representation of the observed space. It discretizes the space based on the Voronoi regions whose centroid is the node created by the Instantaneous Topological Map (ITM) \cite{Jockusch1999} which acts as a self organizing feature map. The adjacent Voronoi regions are connected by Delaunay edges. A transition can exist only between these adjacent regions. A GHMM, at its base, is an HMM with varying number of states and thus, a varying number of transitions. Thus it can be modeled parametrically as $\lambda = (\pi_0, A, b)$; where $A$ is the transition matrix, $\pi_0$ is the initial prior and $b$ is the emission matrix. These parameters can be learnt, based on observed data, using the Incremental Baum-Welch Algorithm \cite{Neal1998} for example, and the model used for inference and prediction.

\begin{figure}[h!]
\centering
\includegraphics[width=\columnwidth]{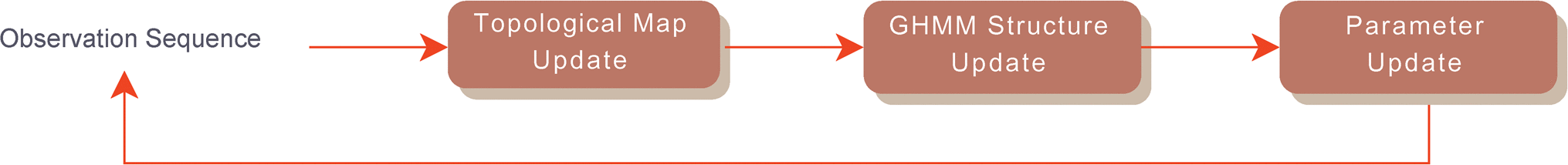}
\caption{Original Growing Hidden Markov Model architecture from \protect\cite{Vasquez2009}.}
\label{fig:original_ghmm}
\end{figure}

A detailed discussion on GHMMs can be found in \cite{Vasquez2009}. Here, we provide a bird's eye view of the different methods involved in its generation and parameter learning. The GHMM is composed of two parts: the ITM and the underlying HMM. Once the topological map has been updated, the underlying HMM can be updated. The states of the HMM correspond to the tuple of the nodes of the ITM and an associated goal. On the map update, nodes and edges in the topology may be added or removed. For every new node added in the ITM, a state with a random initial probability and a random initial transition probability is added to the HMM. When a node is removed, the corresponding states are removed. 
We also make a distinction between the \textit{nodes}, generated by the ITM and the \textit{states} of the GHMM that are generated as a consequence of the nodes. The different steps involved in the generation of a GHMM has been shown in Fig. \ref{fig:original_ghmm}. An incoming observation sequence first updates the topological map. This then updates the structure of the underlying HMM. The sequence is then used to update the parameters of the HMM. 

\subsubsection{Updating the Topological Map}
The topological map discretizes the observed space with nodes and edges. A \textit{node} of the topological map contains 2-Dimensional spatial data associated with its position in the scene. For every observation $O_t$ in a sequence $O_{1:T}$, the nodes of the map are iteratively updated using the ITM method. This method updates the node positions in a direction which minimizes the spatial distance between the node and the associated observations. In this application, these observations are the spatial positions of the pedestrians in the scene. The method may also lead to the addition or removal of nodes from the topology as well as modifying the edges between them.  

An extension to the ITM method was proposed in \cite{Hurtado2015}. This extension allows for considering the flux of people at every node in the topology. The centroids of the ITM are updated in a way as to modify the likelihood function associated with that node. This allows for a flexible observation model, to be able to adapt to different parts of the scenario. 

\subsubsection{Updating the HMM structure}
People move with the intention of reaching a specific goal. Thus, each state in the HMM can be described as $S = (n, p)$ where $n$ is the node and $p$ is the associated goal for the state. An update in the topology results in the corresponding addition or removal of equivalent states. These states are connected if the corresponding nodes are connected on the topological map. Thus the state consists of the 4-Dimensional spatial information $(x_n, y_n, x_p, y_p)$. Also, for every node $n$, there exists an associated Gaussian distribution $G(c_i,\Sigma)$ representing the likelihood of observations at that node, $P(O_t|n)$. In \cite{Vasquez2009}, the covariance $\Sigma$ is considered fixed and unchanging across all the nodes. In \cite{Hurtado2015} however, the Gaussian $G$ varies for each node according to the observations associated with it. 

\subsubsection{Updating the GHMM Parameters}
The GHMM updates probabilities of the prior and the transitions of all its states based on observed trajectories. The original GHMM implementation required a complete trajectory sequence $O_{1:T}$ with the final observation of the trajectory being the goal. The extension in \cite{Hurtado2015} proposes to automatically discover goals in the scene based on the time a person spends at every node. This allows the GHMM algorithm to work with partial trajectories, i.e, when a tracker loses an observed pedestrian due to occlusions, etc. before the end of the trajectory. These can then be coupled with the discovered goals to obtain the corresponding states, $S=(n,p)$. Every observation $O_t$ of the sequence $O_{1:T}$ leads to an update of the ITM and thus the underlying HMM. Once all the observations of the sequence are treated, the Incremental Baum-Welch algorithm is used to update the parameters $\pi$ and $A$ of the model $\lambda$. 

\subsubsection{Motion Prediction}
The HMM, trained on these observation sequences, can then be used to predict future motion. An initial estimation of the person's position is maintained. After every new observation of the position and velocity, the current belief on the position and goal is re-estimated using Bayes' theorem. This belief can be propagated over a finite horizon to estimate the future state of the pedestrian being observed. 

\pagebreak
\subsubsection{Drawbacks} \label{subsec:drawbacks}
While a good method for predicting pedestrian behavior, the current implementations of the GHMM suffers from a few drawbacks: 
\begin{itemize}
	\item \textit{Preset Random Priors}: For every new node $n$ created on the topological map, the corresponding state in the GHMM is initialized with a prior with a random default value, $\pi_n = \pi_0$.   
	\item \textit{Preset Random Transition values}: For every edge added between two nodes $i$ and $j$, the transitions $a_{i,j}$ and $a_{j,i}$ are initialized to a random default value, $a_0$.  
	\item \textit{Rich datasets}: For the ITM to grow a topology covering the entire observed environment, a rich dataset with many well defined observation sequences is required. It also requires these trajectories to be tagged with their corresponding goals. Thus the existing methods cannot be applied for sparse datasets.
\end{itemize}

The contribution of the present work is to extend the above presented state of the art to solve the discussed drawbacks, specifically a way to use the method to solve the problems arising due to sparse, untagged observations in a previously unseen environment. To avoid the problem of preset random priors and transitions arising due to the topological update, we draw upon our previous work \cite{Vasishta2017} which will be expanded below.

\subsection{Natural Vision Based method for generating Potential Cost Maps}\label{subsec:naturalvision}
In the sociological sciences, pedestrian behavior in urban areas has been postulated as being a function of the built environment, i.e, the environment influences the direction and areas of pedestrian movement. There are certain positive and negative \textit{attractors} \cite{Hillier1993} in the environment that push or pull pedestrians in certain directions. This behavior, called Natural Motion, is an extension of Gibson's \textit{Natural Vision} which envisages human behavior as wanting to move in a direction that interests them the most in their field of view \cite{Gibson1979}. In our work, these attractors in the scene are called ``Points of Interest (POIs)''. 

These elements in the scene lead to legal and illegal crossings. Legal pedestrian crossings are those wherein the pedestrian follows only the crosswalk to cross to the other side of the road. Any other path taken to cross to the other side is considered an anomalous crossing and thus ``illegal''. 

In a structured urban environment, for legal crossings to occur, certain assumptions are made \cite{Vasishta2017}: 
\begin{itemize}
	\item The edges of the road repel pedestrians such that their paths are restricted to the side-walk.
	\item A cross-walk acts as a conduit between the two sides of the street and offers no resistance to crossing
	\item The road acts as a barrier for crossing, repelling pedestrians towards the side-walks. 
	\item Static and Dynamic obstacles on the road are repulsive in nature, increasing the resistance of the road and pushing back pedestrians towards side-walks.
	\item Side-walks offer no resistance to pedestrian movement.
	\item Points of Interest are a reason for pedestrians to cross from one side of the street to another. 
\end{itemize}

An illegal crossing occurs when at least one of these assumptions is violated. To model these assumptions, different potential functions are used. 

\subsubsection{Choice of Destinations}
A point of interest is an attractor for the pedestrian. As such, it becomes a destination. Conversely, a destination can also be considered a POI. In an urban environment, one may find many POIs like monuments, places of public interest, public transportation, stores, restaurants, etc., \cite{Helbing1995}. In an observed scene, its viable edges become destinations for pedestrians. 

\subsubsection{Generating the Potential Cost Map} \label{subsubsec:genpf}
Given the positions of POIs in a scene, a potential cost map can be generated using other semantic information extracted from it. The potential cost map is defined as 
\begin{equation}\label{eq:utotal}
\begin{aligned}
\mathbf{U_{total}} = {} & \mathbf{U_{Edge}}+\mathbf{U_{Road}}+\mathbf{U_{Obs}}+\mathbf{U_{POI}}
\end{aligned}
\end{equation}

where $\mathbf{U_{total}}$ in eqn.(\ref{eq:utotal}) is a linear sum of potentials due to the edges of the road, the potential of the road - which is a function of its width -, the potentials due to the obstacles in the scene and the potential values due to the different POIs in the scene. 

The information regarding the width of the street, the positions of the POI and the edges can be extracted by using publicly available data like OpenStreetMaps. Other pertinent data can be extracted from the scene itself, like the positions of the static and dynamic obstacles using a myriad of techniques. 

With this data available, the potential cost map for the observed scene can be generated. 
\section{Theoretical Approach} \label{sec:approach}

Pedestrian behavior on an urban street is a function of the built environment the person is. A good prediction system must be able to take into account these elements in the environment to accurately gauge pedestrian intent so as to provide good situational awareness for the autonomous car.

A novel method for generating a potential cost map to take into account this built environment has been explained in Section \ref{subsec:naturalvision}. The observed scene is semantically labeled as side-walk, cross-walk, road and after identifying the POIs in the scene as well as the static and dynamic obstacles, a resultant potential cost map of the scene is generated \cite{Vasishta2017}. An example of the generated potential cost map is shown in Fig. \ref{fig:itm_itm}.

Structurally, a GHMM is identical to a regular HMM. Thus, the GHMM is generally defined in terms of three variables - $S_t$, $S_{t-1}$ and $O_t$ which are, respectively, state at time $t$, state at time $t-1$ and the observation variable at that time. Using this, the joint probability distribution (JPD) can be defined  \cite{Vasquez2009} as: 

\begin{equation} \label{eqn:jpd_ghmm}
P(S_{t - 1}\ S_{t}\ O_{t}) = \underbrace{P(S_{t - 1})}_{\rm state\,prior} \quad \underbrace{P(S_{t} \vert S_{t - 1})}_{\rm transition \atop probability} \quad \underbrace{P(O_{t} \vert S_{t})}_{\rm observation \atop probability}
\end{equation}

The State Prior is the posterior of the previous time step:
\begin{equation}
P(S_{t - 1}) = P(S_{t - 1} \vert O_{1:t - 1})
\end{equation}
~

\begin{figure}
\centering
\frame{\includegraphics[width=0.6\columnwidth]{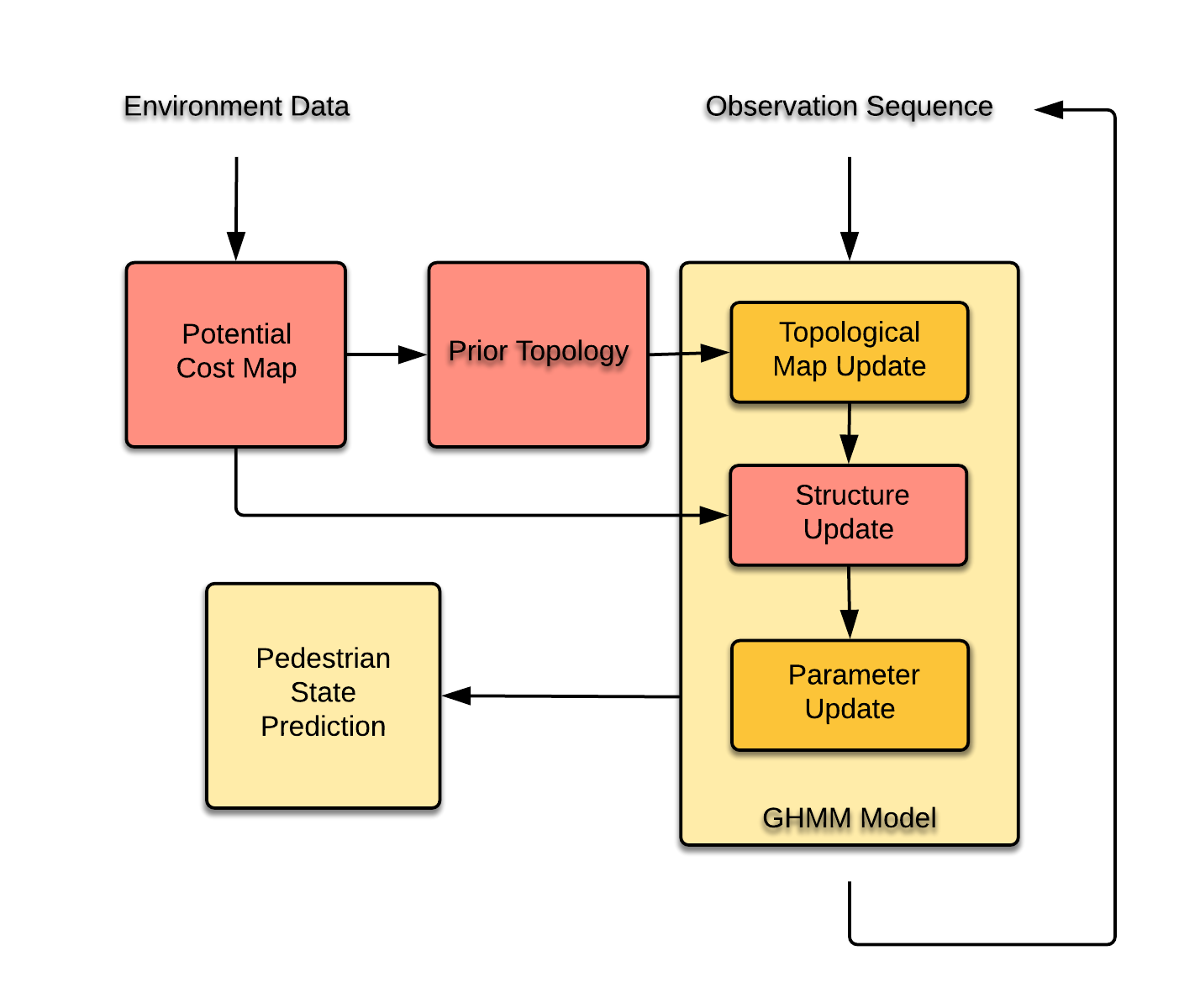}}
\caption{Architecture of the proposed GHMM extension. The red colored blocks denote a divergence from the original GHMM architecture as presented in fig. \ref{fig:original_ghmm}. }
\label{fig:architecture}

\end{figure}

Figure \ref{fig:architecture} shows the architecture of the proposed method. The red blocks show the differences in the implementation compared to the original implementation shown in Fig. \ref{fig:original_ghmm} \cite{Vasquez2009}. The additions are the blocks for generating the potential cost map, which is then used to generate the ``Prior Topology''. This is then subjected to observation sequences, which modify the topology using the ITM, update the GHMM structure and then update the different parameters of the GHMM. The modifications to each of these steps in our implementation is discussed below.

\subsection{Prior Topological Map} \label{subsec:itm_itm}
The potential cost map is generated (Section \ref{subsubsec:genpf}) using environmental data providing a potential cost for each point on the image. From this map, points representing the ground plane of the observed environment is extracted. This ground plane can be estimated using the camera parameters of the observing camera. The points form a grid of interval ``\textit{Insertion Threshold}'' $(\tau)$, which is equal to the one required by the ITM to update the topological map. This set of points also includes all the identified destinations in the scene. 

\begin{figure*}
\centering
	\hfill
    \includegraphics[width=0.45\columnwidth]{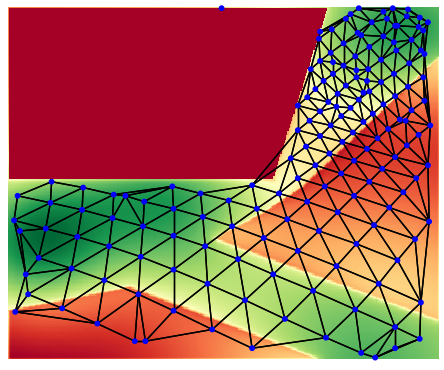}
    \hfill
    \includegraphics[width=0.45\columnwidth]{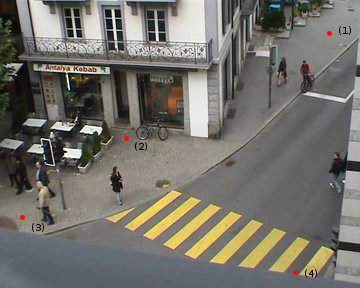}
    \hfill
\caption{Prior Topological Map of the dataset from the Traffic Anomaly Dataset. Figure on the left shows the generated potential cost map and the ``Prior Topology'' of the image from scene on the right.}
\label{fig:itm_itm}
\end{figure*}

These points are the centroids of the Voronoi regions of the area under observation, so as to comply with the requirements of the ITM \cite{Jockusch1999} method. These Voronoi regions, adjacent to each other, can be connected by edges using the Delaunay Triangulation. These edges act as transitions between the states in the Prior Topological Map. Figure \ref{fig:itm_itm} shows a prior topological map at time $t = 0$ on a generated potential cost map. As concurrent observations arrive, this prior topological map can be updated as in \cite{Hurtado2015}.  

\subsection{Structure and Parameter Update of the GHMM.} \label{subsec:learning}
The issues with the current implementations of the GHMMs have been enumerated in Sec. \ref{subsec:drawbacks}. 
This section deals with proposed extensions to mitigate these issues. 

If a connection exists between nodes $i$ and $j$ on the topological map, a transition exists between the states $S_i$ and $S_j$. The transition probability, $P(S_i \mid S_j)$ is dependent on the potential cost of the corresponding nodes of the states $S_i$ and $S_j$. 

A function $f(\cdot)$ is defined such that $f:n_k\mapsto y\in(0,1]$, where $n_k$ is the spatial location of the $k^{th}$ node and $y$ is the resultant potential cost derived from the cost map, which lies between 0 and 1.  

Defining $\alpha = f(n_i)$ and $\beta = f(n_j)$, the transition probability $P(S_i \mid S_j)$ is defined as:

\begin{equation} \label{eqn:transitionfunc}
  P(S_i \mid S_j) =\begin{cases}
               0.8 & \text{if}\ \beta - \alpha > 0\ \land \vert\beta - \alpha\vert > \epsilon \\
               0.5 & \text{if}\  \vert\beta - \alpha\vert \leq \epsilon \\
               0.2 & \text{if}\ \beta - \alpha < 0\ \land \vert\beta - \alpha\vert > \epsilon \\
               0.05 & \text{if}\ \beta - \alpha = 0\ 
            \end{cases}
\end{equation}

In (\ref{eqn:transitionfunc}), one can observe the different values chosen, respectively, for a state transition from a higher potential cost to a lower potential cost; a transition between states of similar cost; a transition from a lower cost state to a higher cost state and finally, a self transition. $\epsilon$ is a threshold value.  

Using this, the following changes can be made to the structure and parameter update of the GHMM:
\begin{enumerate}
	\item When a new node $i$ is added in the topological map, corresponding states $S_i$ are added in the GHMM. These states are initialized with a prior value derived from the potential cost map such that $\pi_i = 1 - f(n_i)$ and $n_{i}$ contains the spatial location of the node. These values will be normalized in the next step obtain probabilities that sum to 1. 
	\item For every state in the GHMM structure, the transitions between all the states are the initialized using eqn. \ref{eqn:transitionfunc}. This initial transition probability is for every state is normalized to add up to 1. On the addition of new states, the same eqn. \ref{eqn:transitionfunc} is used to evaluate the transitions between the new states and existing states. The transitions are then normalized to add up to 1 for each state in the next step.
\end{enumerate}

The different parameters are then updated based on observation sequences as in Hurtado et al. \cite{Hurtado2015} to obtain the GHMM model $\lambda$.

\subsection{Pedestrian State and Goal Prediction} \label{subsec:prediction}
The trained model $\lambda$ can then be used for predicting pedestrian goals and future positions. The state is a tuple of the spatial positions of the node and the goal. A belief is maintained around this state after every time step. After every observation sequence, the belief is recalculated. This is given by eqn. (\ref{eqn:inference}) \cite{Hurtado2015}:  

\begin{equation} \label{eqn:inference}
P(S_{t}\mid O_{1:t})= \frac{1}{\eta}P(O_{t}\vert S_{t})\sum_{S_{t-1}}P(S_{t}\vert S_{t-1})P(S_{t-1}\vert O_{1:t-1}) 
\end{equation}

where $\eta$ is the normalizing constant. 
The beliefs around the current position of the pedestrian and the belief over the goals are given by eqns. (\ref{eqn:pos}) and (\ref{eqn:goal}):

\begin{equation} \label{eqn:pos}
P(n_{t}\mid O_{1:t})= \frac{1}{\eta}\sum_{p}P(S_{t}=(n, p)\mid O_{1:t}) 
\end{equation}

\begin{equation}\label{eqn:goal}
P(p_{t}\mid O_{1:t})= \frac{1}{\eta}\sum_{n}P(S_{t}=(n, p)\mid O_{1:t}) 
\end{equation}

The predictions of the state, H timesteps into the future is given by (\ref{eqn:prediction}) as below. 

\begin{equation} \label{eqn:prediction}
P(S_{t+H}\vert O_{1:t})=\sum_{S_{t+H-1}}P(S_{t+H}\vert S_{t+H-1})P(S_{t+H-1}\vert O_{1:t})
\end{equation}

\section{EXPERIMENTS AND VALIDATION} \label{sec:implementation}
 
The method presented in this work has been tested on the publicly available Traffic Anomaly Detection dataset \cite{Varadarajan2009}. This dataset contains a video which was recorded from a fixed camera in Martigny, Switzerland. The camera overlooks a street in an urban center that contains two POIs - a restaurant, visible in the scene and a commercial store that is unseen in the scene, a crosswalk, sidewalks and observes pedestrians and dynamic obstacles. As such, there are four destinations in the scene - the POIs and the viable edges of the scene.  A shortcoming of this trajectory was the unavailability of many key data that were necessary for the testing of the presented method. For this reason, the video was manually annotated for all observed pedestrian trajectories for positions, velocities, final destination of the trajectory etc.

For this dataset, the observed location is known and thus can be geo-localised. This leads to easy identification of the different POIs that are present in the scenes using resources like OpenStreetMaps which also provides information regarding the width of the street and other parameters required for the generation of the potential cost map. 

To simulate real world settings, pedestrians, as well as dynamic obstacles like cars and buses, were identified using the YOLOv2 framework \cite{redmon2016} tracked with the deepSORT \cite{Wojke2017} algorithm, thereby obtaining pedestrian partial trajectories where a partial trajectory is one where a pedestrian's tracking is lost before the end of the entire trajectory. Of these, 250 partial trajectories were extracted for training our model pedestrian behavior. From the same dataset, 85 unique full trajectories comprising of legal and illegal behaviors on the road were extracted for testing the trained model. A qualitative and quantitative validation of the proposed method is described in sections \ref{subsec:qual} and \ref{subsec:quant}.

\subsection{Qualitative Validation Results} \label{subsec:qual}
The motivation of this work, as mentioned earlier, is to be able to predict pedestrian behavior even with sparse observed data. This implies that the algorithm must be able to anticipate previously unobserved anomalous behavior. Anomalous behavior can be described as being illegal crossings - when pedestrians cross in areas not designated for crossing; on the road, for example. We compare our method for growing the HMM against the one presented in Hurtado et al. \cite{Hurtado2015}.  

\begin{figure*}
     \centering
     Our proposal\vspace{0.025cm}\\
    \includegraphics[width=0.23\textwidth]{images/init_tm.png}
    ~
    \includegraphics[width=0.23\textwidth]{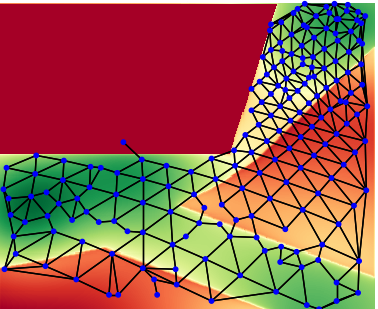}
    ~
    \includegraphics[width=0.23\textwidth]{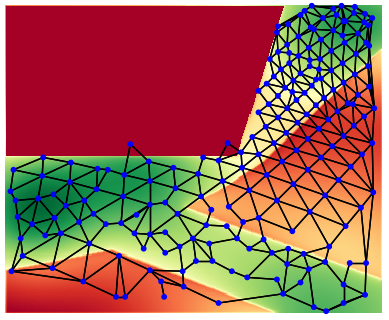}
    ~
    \includegraphics[width=0.23\textwidth]{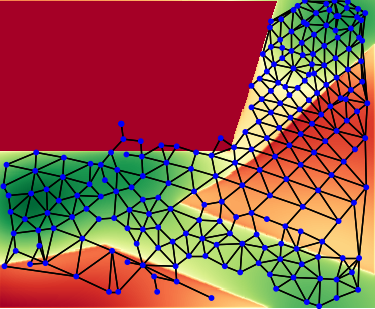}
    \\
    \vspace{0.2cm}
    Previous work from P{\'e}rez-Hurtado et al. \protect\cite{Hurtado2015} \vspace{0.03cm}\\
    \includegraphics[width=0.23\textwidth]{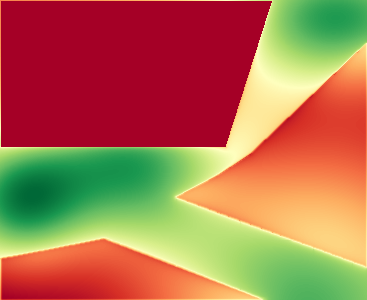}
    ~
    \includegraphics[width=0.23\textwidth]{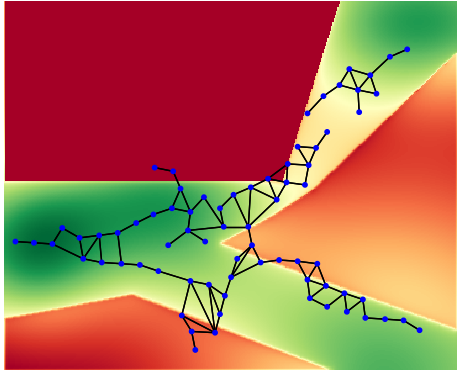}
    ~
    \includegraphics[width=0.23\textwidth]{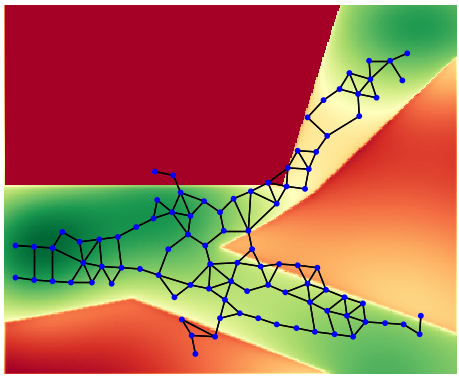}        
    ~
    \includegraphics[width=0.23\textwidth]{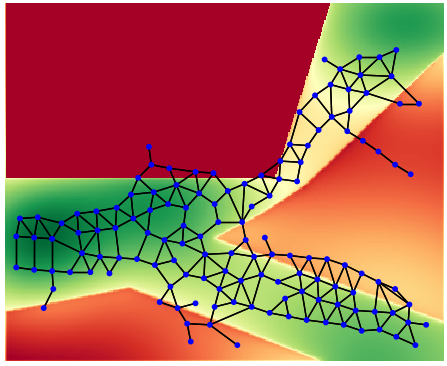}        
    \caption{Comparison between the Topological Map growth between the proposed method (top) and \cite{Hurtado2015} (bottom). Topological map grown at 0, 50, 100 and 250 partial trajectories. }
    \label{fig:itm_dataset}
    
\end{figure*}

Fig. \ref{fig:itm_dataset} shows a qualitative comparison between the proposed method and the current state of the art. It shows the growth of the topology of the GHMM at different number of trajectories. The first image on the left shows the Initial Topology of the environment, before any observed partial trajectories. The second and third images show the growth of the topology at 50 and 100 training trajectories taken from the two datasets. As can be seen, there is a stark contrast in Fig. \ref{fig:itm_dataset}. Even at very few partial trajectories, the environment is sufficiently covered in the proposed approach as to be able to do richer inference of motion and goals compared to the state of the art, allowing for capture of most anomalous behavior on the road. More accurate tracking will lead to a better generation of the topological map. With this trained HMM, pedestrian behavior can then be inferred as presented in eqns. \ref{eqn:inference} and \ref{eqn:prediction}. 

\subsection{Quantitative Validation Results}\label{subsec:quant}

Within the ambit of quantitative validation, we perform long horizon prediction of pedestrian behavior and test the error in the actual position and the predicted position.

The GHMM models were trained for 50 partial trajectories, 100 partial trajectories and 250 partial trajectories. The topological map for the environment at these trajectories are the same as Fig. \ref{fig:itm_dataset}. A test set of 50 complete legal trajectories, i.e., where there is no illegal crossing of the road and another test set of 35 trajectories of illegal behavior.

The test trajectories are used to infer the state of the pedestrian at a horizon of H=75 time steps. The same partial trajectories used for training the proposed model were used for training the GHMM model proposed in Hurtado et al. \cite{Hurtado2015}. Finally, this model was trained with all available testing trajectories and the prediction compared to those from our proposed model with no training data i.e., with only the ``Prior Topology'' and the prior values extracted from the potential cost map.

To test if the proposed method was better at prediction compared to the state of the art, a hypothesis was chosen stating that the prediction was just as good as the state of the art. Errors in prediction at each timestep of every test trajectory were calculated for the stated horizon. A t-test was run on this and p-values for each test trajectory were obtained. These p-values were combined for each class of trajectories, i.e., ``legal'' and ``illegal'', for different number of training trajectories. 
The results of these tests can be seen in Table \ref{table:positionprediction}. As can be seen, the $p$-value is very small for all tested cases. This implies that the proposed method performs significantly better than the state of the art at a horizon of 75 time steps in the future.

\begin{table}[]
\centering
\begin{tabular}{c|c|l|c|l|c|l|l|l|}
\cline{2-9}
\multicolumn{1}{l|}{}                                       & \multicolumn{8}{c|}{\textbf{Number of learning (partial) trajectories}}                                                                                                                         \\ \hline
\multicolumn{1}{|c|}{\multirow{2}{*}{Trajectory type}} & \multicolumn{2}{c|}{50}                     & \multicolumn{2}{c|}{100}                    & \multicolumn{2}{c|}{250}                    & \multicolumn{2}{c|}{0 - Full}               \\ \cline{2-9} 
\multicolumn{1}{|c|}{}                                      & \multicolumn{2}{c|}{p-value}                & \multicolumn{2}{c|}{p-value}                & \multicolumn{2}{c|}{p-value}                & \multicolumn{2}{c|}{p-value}                \\ \hline
\multicolumn{1}{|c|}{Legal}                                 & \multicolumn{2}{c|}{7.023e-16} & \multicolumn{2}{c|}{1.314e-18} & \multicolumn{2}{c|}{2.373e-39} & \multicolumn{2}{l|}{2.172e-61} \\ \hline
\multicolumn{1}{|c|}{Illegal}                               & \multicolumn{2}{c|}{5.602e-33} & \multicolumn{2}{c|}{3.006e-33}  & \multicolumn{2}{c|}{1.178e-29} & \multicolumn{2}{l|}{4.907e-21} \\ \hline
\end{tabular}
\caption{Comparison of prediction accuracy at horizon=75 at varying training levels between the proposed work and \cite{Hurtado2015}.}
\label{table:positionprediction}
\end{table}

\begin{figure*}[h!]
\centering
    \begin{subfigure}[b]{\linewidth}
        \centering
        \includegraphics[width=0.48\textwidth]{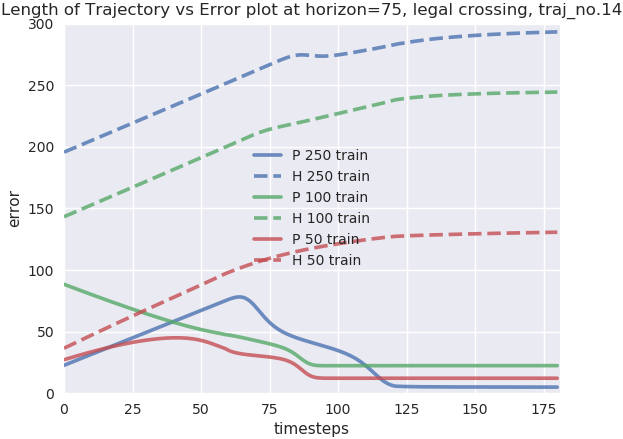}
        \hfill
        \includegraphics[width=0.48\textwidth]{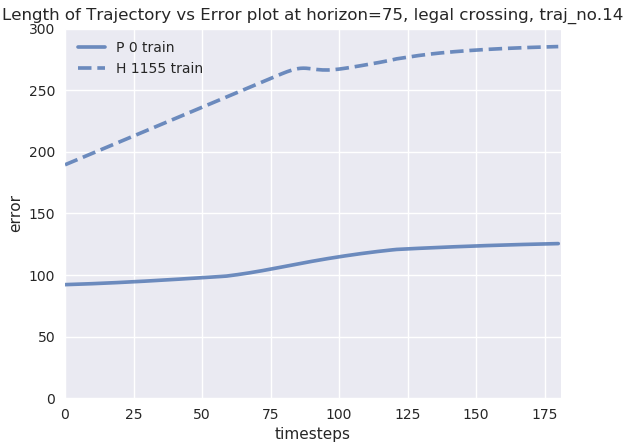}
        \vspace{-5pt}
        \caption{Prediction Error, Legal trajectory}
        \label{subfig:legal}
    \end{subfigure}
    ~ 
    \begin{subfigure}[b]{\linewidth}
        \centering
        \includegraphics[width=0.48\textwidth]{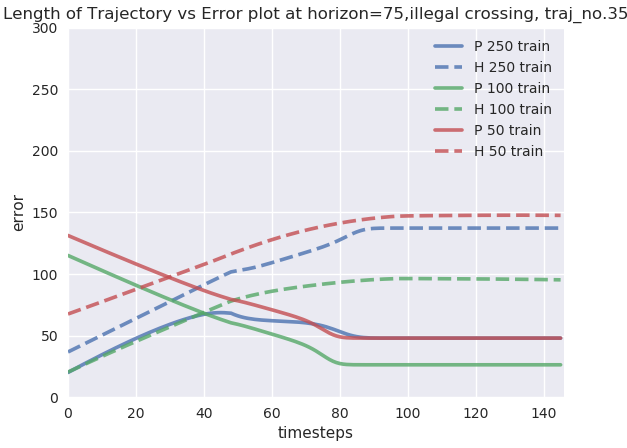}
        \hfill
        \includegraphics[width=0.48\textwidth]{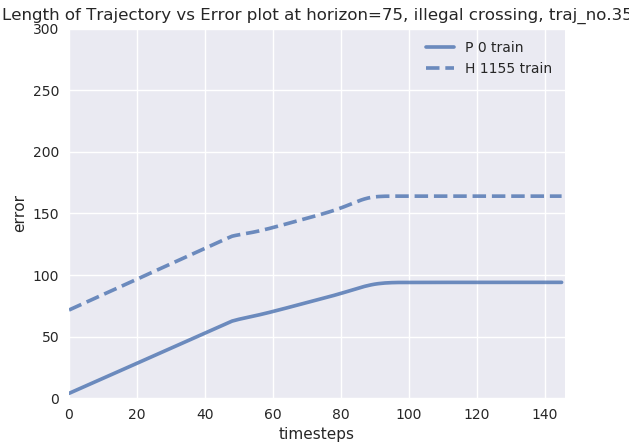}
        \caption{Prediction Error, Illegal trajectory}
        \label{subfig:illegal}
    \end{subfigure}
\caption{Error comparison between the proposed method (P) and \cite{Hurtado2015} (H) for legal and illegal trajectories at horizon=75 with training of 50, 100 and 250 partial trajectories. Figures on the right depict the errors for the same trajectories with a fully trained (H) model vs. the proposed method (P) at 0 training.}
\label{fig:errorcomparison}
\vspace{0.5cm}
\end{figure*}

The last column in Table \ref{table:positionprediction} shows that even with no training, our method performs significantly better than the a fully trained state of the art model.

A secondary analysis on the performance of our method compared to the state of the art is to observe the differences in prediction error on the same trajectory with the models trained on the same training trajectories at different levels which can be seen in Fig. \ref{fig:errorcomparison}.

Figure \ref{subfig:legal} shows the evolution of error over the lifetime of a legal crossing trajectory at different levels of training. As can be seen, the error for all cases in the proposed method quickly reduce to a very small value while that of the state of the art increases over time.   Fig. \ref{subfig:illegal} shows the evolution of error over the lifetime of an illegal crossing trajectory. Though the proposed method starts off predicting with a higher error compared to the state of the art, it quickly reduces to a lower value for the rest of the trajectory. In both the cases, the figures on the right depict the evolution of error of the same trajectory on a fully trained state of the art vs. the proposed method with only the ``Prior Topology''. With increased number of training trajectories, the error reduces at different levels of training as observed in the figures on the left. As can be seen, the proposed method outperforms the state of the art for sample test trajectories in most cases, satisfying the requirement of predicting anomalous behavior. 

\section{CONCLUSION} \label{sec:conclusion}

This work demonstrates an extension to the Growing Hidden Markov Model method, drawing on previously published work based on \textit{Natural Vision} \cite{Vasishta2017}. This paper first discusses the shortcomings of the state of the art in GHMM methods. Following this, it proposes an extension to overcome these shortcomings. This extension allows for the generation of a ``Prior Topology'' based on the semantics of the observed environment. This ``Prior Topology'' is then trained by the use of the Instantaneous Topological Mapping method \cite{Jockusch1999} which results in a self organized mapping of the environment. The underlying HMM is trained using this map and previously observed partial trajectories of pedestrians. It has also been demonstrated here that the current method is more robust and is more accurate at prediction pedestrian positions over a long horizon compared to the state of the art. Indeed, it is demonstrated that an untrained model, i.e., a model that has not observed any pedestrian trajectories and is composed of only the ``Prior Topology'', performs, in some cases, better than a fully trained model of the state of the art and, at worst, is at par with the state of the art \cite{Hurtado2015}. 

Pedestrian behavior in urban areas change based on dynamic obstacles preset in the environment. For example, a car stopping on the street for a short period of time causes a change in the route choice of the pedestrian. Future work will focus on incorporating this hypothesis into the model. Further, it is envisaged to implement this method for use from the perspective of a moving car in urban areas. This would be two fold - to generate the potential cost map from the moving car based on identified and observed semantics and then using the current method to predict observed pedestrians' behavior. This would facilitate safer navigation in pedestrian-heavy environments for an autonomous vehicle.  

\section*{ACKNOWLEDGMENT}

These researches have been conducted within the VALET project, funded by the French Ministry of Education and Research and the French National Research Agency (ANR-15-CE22-0013-02). The author also wishes to thank Dr. Jilles Debangoye and David Sierra Gonzalez for their fruitful inputs and discussions that led to this paper.

\bibliographystyle{IEEEtran}
\bibliography{biblio/icarcv18}

\begin{thebibliography}{10}
\providecommand{\url}[1]{#1}
\csname url@samestyle\endcsname
\providecommand{\newblock}{\relax}
\providecommand{\bibinfo}[2]{#2}
\providecommand{\BIBentrySTDinterwordspacing}{\spaceskip=0pt\relax}
\providecommand{\BIBentryALTinterwordstretchfactor}{4}
\providecommand{\BIBentryALTinterwordspacing}{\spaceskip=\fontdimen2\font plus
\BIBentryALTinterwordstretchfactor\fontdimen3\font minus
  \fontdimen4\font\relax}
\providecommand{\BIBforeignlanguage}[2]{{%
\expandafter\ifx\csname l@#1\endcsname\relax
\typeout{** WARNING: IEEEtran.bst: No hyphenation pattern has been}%
\typeout{** loaded for the language `#1'. Using the pattern for}%
\typeout{** the default language instead.}%
\else
\language=\csname l@#1\endcsname
\fi
#2}}
\providecommand{\BIBdecl}{\relax}
\BIBdecl

\bibitem{Endsley1995}
M.~R. Endsley, ``Toward a theory of situation awareness in dynamic systems,''
  \emph{Human factors}, vol.~37, no.~1, pp. 32--64, 1995.

\bibitem{Ziebart2009}
B.~D. Ziebart, N.~Ratliff, G.~Gallagher, C.~Mertz, K.~Peterson, J.~A. Bagnell,
  M.~Hebert, A.~K. Dey, and S.~Srinivasa, ``Planning-based prediction for
  pedestrians,'' in \emph{IEEE/RSJ International Conference on Intelligent
  Robots and Systems (IROS)}.\hskip 1em plus 0.5em minus 0.4em\relax IEEE,
  2009, pp. 3931--3936.

\bibitem{Kitani2012}
K.~M. Kitani, B.~D. Ziebart, J.~A. Bagnell, and M.~Hebert, ``Activity
  forecasting,'' in \emph{European Conference on Computer Vision}.\hskip 1em
  plus 0.5em minus 0.4em\relax Springer, 2012, pp. 201--214.

\bibitem{Bandyopadhyay2013}
T.~Bandyopadhyay, C.~Z. Jie, D.~Hsu, H.~Marcelo, A.~Jr, D.~Rus, and
  E.~Frazzoli, ``{Intention-Aware Pedestrian Avoidance},'' \emph{The 13th
  International Symposium on Experimental Robotics}, pp. 963--977, 2013.

\bibitem{Yamaguchi2011}
K.~Yamaguchi, A.~C. Berg, L.~E. Ortiz, and T.~L. Berg, ``Who are you with and
  where are you going?'' in \emph{IEEE Conference on Computer Vision and
  Pattern Recognition (CVPR)}.\hskip 1em plus 0.5em minus 0.4em\relax IEEE,
  2011, pp. 1345--1352.

\bibitem{Vasquez2009}
D.~Vasquez, T.~Fraichard, and C.~Laugier, ``Incremental {L}earning of
  {S}tatistical {M}otion {P}atterns with {G}rowing {H}idden {M}arkov
  {M}odels,'' \emph{IEEE Transactions on Intelligent Transportation Systems},
  vol.~10, no.~3, pp. 403--416, 2009.

\bibitem{Hurtado2015}
I.~P{\'e}rez-Hurtado, J.~Capit{\'a}n, F.~Caballero, and L.~Merino, ``An
  extension of {GHMM}s for environments with occlusions and automatic goal
  discovery for person trajectory prediction,'' in \emph{European Conference on
  Mobile Robots (ECMR)}.\hskip 1em plus 0.5em minus 0.4em\relax IEEE, 2015, pp.
  1--7.

\bibitem{Elfring2014}
J.~Elfring, R.~{Van De Molengraft}, and M.~Steinbuch, ``{Learning intentions
  for improved human motion prediction},'' \emph{Robotics and Autonomous
  Systems}, vol.~62, no.~4, pp. 591--602, 2014.

\bibitem{Helbing1995}
D.~Helbing and P.~Molnar, ``Social force model for pedestrian dynamics,''
  \emph{Physical review E}, vol.~51, no.~5, p. 4282, 1995.

\bibitem{Gibson1979}
J.~J. Gibson, ``The ecological approach to visual perception.'' 1979.

\bibitem{Vasishta2017}
P.~Vasishta, D.~Vaufreydaz, and A.~Spalanzani, ``Natural {V}ision {B}ased
  {M}ethod for {P}redicting {P}edestrian {B}ehaviour in {U}rban
  {E}nvironments,'' in \emph{IEEE 20th International Conference on Intelligent
  Transportation Systems}, 2017.

\bibitem{Jockusch1999}
J.~Jockusch and H.~Ritter, ``An {I}nstantaneous {T}opological {M}apping {M}odel
  for {C}orrelated {S}timuli,'' in \emph{International Joint Conference on
  Neural Networks, (IJCNN)}, vol.~1, 1999, pp. 529--534 vol.1.

\bibitem{Neal1998}
R.~M. Neal and G.~E. Hinton, ``A view of the {EM} algorithm that justifies
  incremental, sparse, and other variants,'' in \emph{Learning in graphical
  models}.\hskip 1em plus 0.5em minus 0.4em\relax Springer, 1998, pp. 355--368.

\bibitem{Hillier1993}
B.~Hillier, A.~Penn, J.~Hanson, T.~Grajewski, and J.~Xu, ``Natural movement:
  or, configuration and attraction in urban pedestrian movement,''
  \emph{Environment and Planning B: planning and design}, vol.~20, no.~1, pp.
  29--66, 1993.

\bibitem{Varadarajan2009}
J.~Varadarajan and J.-M. Odobez, ``Topic models for scene analysis and
  abnormality detection,'' in \emph{IEEE 12th International Conference on
  Computer Vision Workshops (ICCV Workshops)}.\hskip 1em plus 0.5em minus
  0.4em\relax IEEE, 2009, pp. 1338--1345.

\bibitem{redmon2016}
J.~Redmon and A.~Farhadi, ``Yolo9000: Better, faster, stronger,'' \emph{arXiv
  preprint arXiv:1612.08242}, 2016.

\bibitem{Wojke2017}
N.~Wojke, A.~Bewley, and D.~Paulus, ``{Simple Online and Realtime Tracking with
  a Deep Association Metric},'' \emph{arXiv preprint arXiv:1703.07402}, 2017.

\end{thebibliography}

\end{document}